\pdfoutput=1
\documentclass[11pt]{article}

\usepackage{naacl2021}
\usepackage{times}
\usepackage{latexsym}

\usepackage[T1]{fontenc}
\usepackage[utf8]{inputenc}

\usepackage{microtype}
\usepackage{mathtools}

\usepackage{latexsym}
\usepackage{wasysym}
\usepackage{cancel}
\usepackage{amsfonts}
\usepackage{soul}
\usepackage{graphicx}
\usepackage{xcolor}
\usepackage{caption}
\usepackage{subcaption}
\usepackage{hyperref}
\title{UDALM: Unsupervised Domain Adaptation through Language Modeling}

\author{Constantinos Karouzos\textsuperscript{1},
        Georgios Paraskevopoulos\textsuperscript{1,4}, 
        Alexandros Potamianos\textsuperscript{1,2,3} \\
\\
  \textsuperscript{1} School of ECE, National Technical University of Athens, Athens, Greece \\
  \textsuperscript{2} Signal Analysis and Interpretation Laboratory (SAIL), USC, Los Angeles, CA, USA \\
  \textsuperscript{3} Behavioral Signal Technologies, Los Angeles, CA, USA \\
    \textsuperscript{4} Institute for Language and Speech Processing, Athena Research Center, Athens, Greece \\
  \texttt{ckarouzos@gmail.com},\texttt{ geopar@central.ntua.gr},\texttt{ potam@central.ntua.gr}}

\begin{document}
\maketitle
\begin{abstract}
In this work we explore Unsupervised Domain Adaptation (UDA) of pretrained language models for downstream tasks.
We introduce UDALM, a fine-tuning procedure, using a mixed classification and Masked Language Model loss, that can adapt to the target domain distribution in a robust and sample efficient manner.
Our experiments show that performance of models trained with the mixed loss scales with the amount of available target data and the mixed loss can be effectively used as a stopping criterion during UDA training.
Furthermore, we discuss the relationship between A-distance and the target error and explore some limitations of the Domain Adversarial Training approach.
Our method is evaluated on twelve domain pairs of the Amazon Reviews Sentiment dataset, yielding $91.74\%$ accuracy, which is an $1.11\%$ absolute improvement over the state-of-the-art.
\end{abstract}

\section{Introduction}

Deep architectures have achieved state-of-the-art results in a variety of machine learning tasks.
However, real world deployments of machine learning systems often operate under domain shift, which leads to performance degradation.
This introduces the need for adaptation techniques, where a model is trained with data from a specific domain, and then can be optimized for use in new settings.
Efficient techniques for model re-usability can lead to faster and cheaper development of machine learning applications and facilitate their wider adoption.
Especially techniques for Unsupervised Domain Adaptation (UDA) can have high real world impact, because they do not rely on expensive and time-consuming annotation processes to collect labeled data for domain-specific supervised training, further streamlining the process.
% kai kamia papatza gia environment / democratization

UDA approaches in the literature can be grouped in three major categories, namely pseudo-labeling techniques (e.g. \citealp{yarowsky-1995-unsupervised, zhou2005tri}), domain adversarial training (e.g. \citealp{ganin2016domain}) and pivot-based approaches (e.g. \citealp{blitzer-etal-2006-domain, pan2010cross}). Pseudo-labeling approaches use a model trained on the source labeled data to produce pseudo-labels for unlabeled target data and then train a model for the target domain in a supervised manner. Domain adversarial training aims to learn a domain-independent mapping for input samples by adding an adversarial cost during model training, that minimizes the distance between the source and target domain distributions. Pivot-based approaches aim to select domain-invariant features (pivots) and use them as a basis for cross-domain mapping.
This work does not fall under any of these categories, rather we aim to optimize the fine-tuning procedure of pretrained language models (LMs) for learning under domain-shift.

Transfer learning from language models pretrained in massive corpora \citep{howard-ruder-2018-universal, devlin-etal-2019-bert, yang2019xlnet, liu2019roberta,brown2020language} has yielded significant improvements across a wide variety of NLP tasks, even when small amounts of data are used for fine-tuning. 
Fine-tuning a pretrained model is a straightforward framework for adaptation to target tasks and new domains, when labeled data are available.
However, optimizing the fine-tuning process in UDA scenarios, where only labeled out-of-domain and unlabeled in-domain data are available is challenging.

In this work, we propose UDALM, a fine-tuning method for BERT  \citep{devlin-etal-2019-bert} in order to address the UDA problem. Our method is based on simultaneously learning the task from labeled data in the source distribution, while adapting to the language in the target distribution using multitask learning. The key idea of our method is that by simultaneously minimizing a task-specific loss on the source data and a language modeling loss on the target data during fine-tuning, the model will be able to adapt to the language of the target domain, while learning the supervised task from the available labeled data.

Our key contributions are: (a) We introduce UDALM, a novel, simple and robust unsupervised domain adaptation procedure for downstream BERT models based on multitask learning, (b) we achieve state-of-the-art results for the Amazon reviews benchmark dataset, surpassing more complicated approaches and (c) we explore how A-distance and the target error are related and conclude with some remarks on domain adversarial training, based on theoretical concepts and our empirical observations. Our code and models are publicly available\footnote{\href{https://github.com/ckarouzos/slp_daptmlm}{https://github.com/ckarouzos/slp\_daptmlm}}.

\section{Related Work}

Traditionally, UDA has been performed using pseudo-labeling approaches. Pseudo-labeling techniques are semi-supervised algorithms that either use the same model (self-training) \citep{yarowsky-1995-unsupervised, mcclosky-etal-2006-reranking, abney2007semisupervised} or multiple ensembles of models (tri-training) \citep{zhou2005tri, sogaard-2010-simple} in order to produce pseudo-labels for the target unlabeled data. \citet{saito2017asymmetric} proposed an asymmetric tri-training approach. \citet{ruder-plank-2018-strong} introduced a multi-task tri-training method. \citet{rotman-reichart-2019-deep} and \citet{lim2020semi}  study pseudo-labeling with contextualized word representations. \citet{ye_feature_2020} combine self-training with XLM-R \citep{conneau-etal-2020-unsupervised} to reduce the produced label noise and propose CFd, class aware feature self-distillation.

Another line of UDA research includes pivot-based methods, focusing on extracting cross-domain features. Structural Correspondence Learning (SCL) \citep{blitzer-etal-2006-domain} and Spectral Feature Alignment \citep{pan2010cross} aim to find domain-invariant features (pivots) to learn a mapping between two domain distributions. \citet{ziser-reichart-2017-neural, ziser-reichart-2018-pivot, ziser-reichart-2019-task} combine SCL with neural network architectures and language modeling. \citet{miller-2019-simplified} propose to jointly learn the task and pivots. \citet{li2018hierarchical} learn pivots with hierarchical attention networks. Pivot-based methods have also been used in conjunction with BERT \citep{ben2020perl}.

Domain adversarial training is a dominant approach for UDA \citep{ramponi-and-plank-2020-neural}, inspired by the theory for learning from different domains introduced in \citet{ben2007analysis, ben2010theory}. \citet{ganin2016domain, ganinlempitsky2015} propose to learn a task while not being able to distinguish if samples come from the source or the target distribution, through use of an adversarial cost. This approach has been adopted for a diverse set of problems, e.g. sentiment analysis, tweet classification and universal dependency parsing \citep{li-etal-2018-whats, alam-etal-2018-domain, sato-etal-2017-adversarial}.  \citet{du-etal-2020-adversarial} pose domain adversarial training in the context of BERT models.  \citet{zhao2018adversarial} propose multi-source domain adversarial networks. \citet{guo-etal-2018-multi} propose a mixture-of-experts approach for multi-source UDA. \citet{guo2020multi} explore distance measures as additional losses and use them to construct dynamic multi-armed bandit controller for the source domains. \citet{shen2017wasserstein} learn domain invariant features via Wasserstein distance.  \citet{bousmalis2016domain} introduce domain seperation networks with private and shared encoders. 

Unsupervised pretraining on domain-specific corpora can be an effective adaptation process.
For example BioBERT \citep{lee2020biobert} and SciBERT \citep{beltagy-etal-2019-scibert} are specialized BERT variants, where pretraining is extended on large amounts of biomedical and scientific corpora respectively. \citet{sun2019fine} propose continuing the pretraining of BERT with target domain data and multitask learning using relevant tasks for BERT fine-tuning. \citet{xu-etal-2019-bert} introduce a review reading comprehension task and a post-training approach for BERT with an auxiliary loss on a question-answering task. Continuing pretraining on multiple phases, from general to domain specific (DAPT) and task specific data (TAPT), further improves performance of pretrained language models, as shown by \citet{gururangan-etal-2020-dont}. \citet{han-eisenstein-2019-unsupervised} propose AdaptaBERT, which includes a second phase of unsupervised pretraining, in order to use BERT in a unsupervised domain adaptation context.

Recent works have highlighted the merits of using Language Modeling as an auxiliary task during fine-tuning. \citet{chronopoulou-etal-2019-embarrassingly} use an auxiliary LM loss to avoid catastrophic forgetting in transfer learning and \citet{jia-etal-2019-cross} adopt this approach for cross-domain named-entity recognition. We draw inspiration from these approaches and utilize auxiliary Language Modeling for UDA.

\begin{figure*}
\centering
\begin{subfigure}[t]{0.84\textwidth}
\centering
\includegraphics[width=\textwidth]{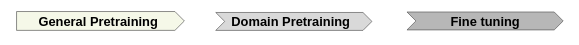}
\end{subfigure}
\begin{subfigure}[t]{0.27\textwidth}
\centering
\includegraphics[width=\textwidth]{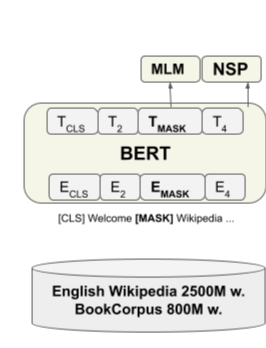}
\caption{}
\label{fig:GPT}
\end{subfigure}
\begin{subfigure}[t]{0.27\textwidth}
\centering
\includegraphics[width=\textwidth]{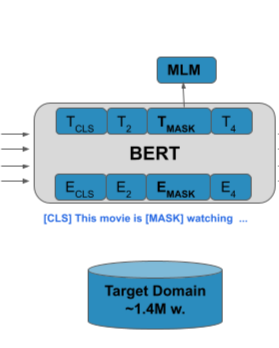}
\caption{}
\label{fig:DPT}
\end{subfigure}
\begin{subfigure}[t]{0.3\textwidth}
\centering
\includegraphics[width=\textwidth]{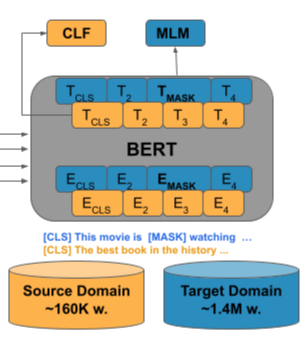}
\caption{}
\label{fig:FT}
\end{subfigure}
    
    \caption{(a) BERT \citep{devlin-etal-2019-bert} is pretrained on English Wikipedia and BookCorpus with the Masked Language Modeling (MLM) and the Next Sentence Prediction (NSP) tasks. (b) We continue the pretraining of BERT on unlabeled target domain data using the MLM task. (c) We train a task classifier with source domain labeled data, while we keep the MLM objective on unlabeled target domain data.}
    \label{fig:figure}
\end{figure*}

\section{Problem Definition}
Let \(X\) be the input space and \(Y\) the set of labels. For binary classification tasks \(Y=\{0,1\}\). In domain adaptation there are two different distributions over \(X \times Y\), called the source domain \(D_S\) and the target domain \(D_T\). In the unsupervised setting labels are provided for samples drawn from \(D_S\), while samples drawn from \(D_T\) are unlabeled. The goal is to train a model that performs well on samples drawn from the target distribution \(D_T\). This is summarized in Eq.~\ref{problem_setting}. 
\begin{equation}\label{problem_setting}
\begin{split}
 &	S = {(x_i, y_i)}_{i=1}^n  \sim (D_S)^n \\
 &  T ={(x_i)}_{i=n+1}^{n+m} \sim (D_T^X)^{m} 
\end{split}
\end{equation}
where \(D_T^X\) is the marginal distribution of \(D_T\) over \(X\), \(n\) is the number of samples from the source domain and \(m\) is the number of samples from the target domain.

\section{Proposed Method}
\label{sec:proposed}

Fig.~\ref{fig:figure} gives an overview of the proposed Unsupervised Domain Adaptation through Language Modeling (UDALM). Starting from a model that is pretrained in general corpora (Fig.~\ref{fig:GPT}), we keep pretraining it on target domain data using the masked language modeling task (Fig.~\ref{fig:DPT}). On the final fine-tuning step (Fig.~\ref{fig:FT}) we update the model weights using both a classification loss on the labeled source data and Masked Language Modeling loss on the unlabeled target data.

In Fig.~\ref{fig:GPT} we see the BERT general pretraining phase. BERT \citep{devlin-etal-2019-bert} is based on the Transformer architecture \cite{vaswani2017attention}. 
% It is trained as a Masked Language Model (MLM), by predicting masked words in the input. 
During BERT pretraining, input tokens are randomly selected to be masked. 
BERT is trained using the Masked Language Modeling (MLM) objective, which consists of predicting the most probable tokens for the masked positions. 
Additionally it uses a Next Sentence Prediction (NSP) loss, which classifies whether the pair of input sentences are continuous or not.
If a labeled dataset is available, a pretrained BERT model can be fine-tuned for the downstream task in a supervised manner with the addition of an output layer. 

In Fig.~\ref{fig:DPT} we initialize a model using the weights of a generally pretrained BERT and continue pretraining on an unsupervised set of in-domain data, in order to adapt to the target domain. This step does not require use of supervised data, since we use the MLM objective.

For the final fine-tuning step, shown in Fig.~\ref{fig:FT} we perform supervised fine-tuning on the source data, while we keep the MLM objective on the target data as an auxiliary task.
Following standard practice, we use the \texttt{[CLS]} token representation for classification. The classifier consists of a single feed-forward layer.

During this procedure the model learns the task through the classification objective using the labeled source domain samples, and simultaneously it adapts to the target domain data through the MLM objective.
The model is trained on the source domain labeled data for the classification task and target domain unlabeled data for the masked language modeling task. We mask only the target domain data. During training we interleave source and target data and feed them to the BERT encoder. Features extracted from the source data are then used for classification, while target features are used for Masked Language Modeling. 

The mixed loss used for the fine-tuning step, is the sum of the classification loss \(L_{CLF}\) and the auxiliary MLM loss \(L_{MLM}\). 
\(L_{CLF}\) is a cross-entropy loss, calculated on labeled examples from source domain, while \(L_{MLM}\) is used to predict masked tokens for unlabeled examples from target domain.
We train the model over mixed batches, that include both source and target data, used for the respective tasks. 
The mixed loss is presented in Eq.~\ref{loss}:
\begin{equation}
\label{loss}
    L(\textbf{s},\textbf{t}) = \lambda L_{CLF}(\textbf{s}) + (1-\lambda) L_{MLM}(\textbf{t})
\end{equation}
We process \(n\) labeled source samples \(\textbf{s} \sim D_S\) and \(m\) unlabeled target samples \(\textbf{t} \sim D_T\) on a batch. The weighting factor  \(\lambda\) is selected as the ratio of labeled source data over the sum of labeled source and unlabeled target data, as stated in Eq.~\ref{lamda}:
\begin{equation}
\label{lamda}
    \lambda = \frac{n}{n+m} 
\end{equation}

\section{Experiments}
\subsection{Dataset}

We evaluate UDALM on the Amazon reviews multi-domain sentiment dataset \citep{blitzer-etal-2007-biographies}, a standard benchmark dataset for domain adaptation. Reviews with one or two stars are labeled as negative, while reviews with four or five stars are labeled as positive. 
The dataset contains reviews on four product domains: \textit{Books} (B), \textit{DVDs} (D), \textit{Electronics} (E) and \textit{Kitchen appliances} (K), yielding $12$ adaptation scenarios of source-target domain pairs. 
Balanced sets of $2000$ labeled reviews are available for each domain.
We use $20000$ (randomly selected) unlabeled reviews for (B), (D) and (E). For (K) $17805$ unlabeled reviews are available.
For each of the $12$ adaptation scenarios we use  $20\%$ of both labeled source and unlabeled target data for validation, while labeled target data are used for testing exclusively and are not seen during training or validation. 
% When a domain is considered as target for a particular adaptation scenario, all the available labeled data from this domain are used for testing.

\begin{table*}[ht]
\centering
\small
\begin{tabular}{||l| c c c | c c c c||}
\hline
                    & R-PERL & DAAT & p+CFd & SO BERT & DAT BERT & DPT BERT & UDALM \\
\hline
    $B \rightarrow D$ & $87.8$ & $90.9$ & $87.7$ & $89.51 \pm 0.76$ & $87.31 \pm 2.14$ & $90.49 \pm 0.38$ & $\mathbf{90.97} \pm 0.22$ \\
    $B \rightarrow E$ & $87.2$ & $88.9$ & $91.3$ & $90.51 \pm 0.51$ & $86.91 \pm 2.71$ & $90.38 \pm 1.59$ & $\mathbf{91.69} \pm 0.31$ \\
    $B \rightarrow K$ & $90.2$ & $88.0$ & $92.5$ & $91.75 \pm 0.28$ & $90.59 \pm 1.17$ & $92.66 \pm 0.43$ & $\mathbf{93.21} \pm 0.22$ \\
    $D \rightarrow B$ & 85.6 & 89.7 & \textbf{91.5} & $90.26 \pm 0.64$ & $86.30 \pm 3.10$ & $91.02 \pm 0.75$ & $91.00 \pm 0.42$ \\
    $D \rightarrow E$ & 89.3 & 90.1 & 91.6 & $88.71 \pm 1.48$ & $87.85 \pm 1.24$ & $91.03 \pm 0.82$ & $\mathbf{92.30} \pm 0.47$ \\
    $D \rightarrow K$ & 90.4 & 88.8 & 92.5 & $91.22 \pm 0.69$ & $89.95 \pm 1.53$ & $92.30 \pm 0.42$ & $\mathbf{93.66} \pm 0.37$ \\
    $E \rightarrow B$ & 90.2 & 89.6 & 88.7 & $87.96 \pm 0.89$ & $85.65 \pm 1.91$ & $88.52 \pm 0.55$ & $\mathbf{90.61} \pm 0.30$ \\
    $E \rightarrow D$ & 84.8 & \textbf{89.3} & 88.2 & $87.37 \pm 0.64$ & $83.99 \pm 1.31$ & $87.85 \pm 0.47$ & $88.83 \pm 0.61$ \\
    $E \rightarrow K$ & 91.2 & 91.7 & 93.6 & $93.30 \pm 0.50$ & $92.45 \pm 1.35$ & $94.39 \pm 0.72$ & $\mathbf{94.43} \pm 0.24$ \\
    $K \rightarrow B$ & 83.0 & \textbf{90.8} & 89.8 & $88.15 \pm 0.64$ & $85.07 \pm 1.03$ & $88.83 \pm 0.81$ & $90.29 \pm 0.51$ \\
    $K \rightarrow D$ & 85.6 & \textbf{90.5} & 87.8 & $87.23 \pm 0.49$ & $84.11 \pm 0.62$ & $88.52 \pm 0.69$ & $89.54 \pm 0.59$ \\
    $K \rightarrow E$ & 91.2 & 93.2 & 92.6 & $93.23 \pm 0.34$ & $92.07 \pm 0.24$ & $93.42 \pm 0.40$ & $\mathbf{94.34} \pm 0.26$ \\
    \hline
    Average & 87.50 & 90.12 & 90.63 & $89.93 \pm 0.65$ & $87.68 \pm 1.53$ & $90.78 \pm 0.67$ & $\mathbf{91.74} \pm 0.38$  \\
    \hline
\end{tabular}
\caption{Accuracy of unsupervised domain adaptation on twelve domain pairs of Amazon Reviews Multi Domain Sentiment Dataset.}
\label{tab:results}
\end{table*}

\subsection{Implementation Details}

We use \(BERT_{BASE}\) (uncased) as the Language Model on which we apply domain pretraining. The \(BERT_{BASE}\) original English model is a 12-layer, 768-hidden, 12-heads, 110M parameter transformer architecture, trained on the BookCorpus with 800M words and a version of the English Wikipedia with 2500M words.
We convert source and target sentences to WordPieces \citep{wu2016google}. For target sentences we randomly mask \(15\%\) of WordPiece tokens, as in \citep{devlin-etal-2019-bert}. If a token in a specific position is selected to be masked \(80\%\) of the time is replaced with a \texttt{[MASK]} token, \(10\%\) of the time with a random token and \(10\%\) of the time remains unchanged.

The maximum sequence length is set to 512 by truncation of inputs. During domain pretraining we train with batch size of 8 for 3 epochs (2 hours on two GTX-1080Ti cards).   
During the final fine-tuning step of UDALM we train with batch size 36, consisting of $n=1$ source sub-batch of $4$ samples and $m=8$ target sub-batches of $4$ samples each.
We update parameters after every $5$ accumulated sub-batches. We train for $10$ epochs with early stopping on the mixed loss in Eq.~\ref{loss}. 
For all experiments we use AdamW optimizer \cite{loshchilov2018decoupled} with learning rate $10^{-5}$.
Each adaptation scenario requires one hour on one GTX-1080Ti.
For the domain adversarial experiments we set  \(\lambda_{d}=0.01\) in Eq.~\ref{eq:datobj} \footnote{We also manually experimented with $\lambda_d=1$ and $lambda_d=0.1$, and a sigmoid schedule for \(\lambda_{d}\). We report best results.} and train for $10$ epochs.
Models are developed with PyTorch \citep{pytorch} and HuggingFace Transformers \citep{wolf2019huggingface}.

\subsection{Baselines - Compared methods}

We select three state-of-the-art methods for comparison. Each of the selected methods represents a different line of UDA research, namely domain adversarial training \textbf{BERT-DAAT} \citep{du-etal-2020-adversarial}, self-training XLM-R based \textbf{p+CFd} \citep{ye_feature_2020} and pivot-based \textbf{R-PERL} \citep{ben2020perl}.
We report results for the following settings with BERT models:
% \\

\noindent \textbf{Source only (SO)}: We fine-tune BERT on source domain labeled data, without using target data. 
% \\

\noindent \textbf{Domain Pretraining (DPT)}: We use the target domain unlabeled data in order to continue pretraining of BERT with MLM loss (as in Fig.~\ref{fig:DPT}) and then fine-tune the resulting model on source domain labeled data.
% \\

\noindent \textbf{Domain Adversarial (DAT)}: Domain Adversarial Training with BERT. Starting from the domain pretrained BERT (as in Fig.~\ref{fig:DPT}), we then fine-tune the model with domain adversarial training as in \citet{ganin2016domain}. 
For a BERT model with parameters \(\theta\), with \(L_{CLF}\) being a cross-entropy loss for supervised task prediction, \(L_{ADV}\) being a cross-entropy loss for domain prediction and \(\lambda_{d}\) being a weighting factor, domain adversarial training consists of the minimization criterion described in Eq.~\ref{eq:datobj}.
\vspace*{-1mm}
\begin{equation}
\label{eq:datobj}
    \min_{\theta} L_{CLF}(\theta;D_S) - \lambda_{d} L_{ADV}(\theta; D_S, D_T)
\end{equation}
\vspace*{-4mm}

\noindent\textbf{UDALM}: The proposed method, where we fine-tune the model created in the domain pretraining step using the mixed loss in Eq.~\ref{loss}.

\section{Experimental Results}
\subsection{Comparison to state-of-the-art}

We present results for all 12 domain adaptation settings in Table~\ref{tab:results}. 
Results for SO BERT, DAT BERT, DPT BERT and UDALM are averaged over five runs and we include standard deviations
The last line of Table~\ref{tab:results} contains the macro-averaged accuracy and deviations over all domain pairs. UDALM surpasses all other techniques, yielding an absolute improvement of \(1.81\%\) over the SO BERT baseline. For fair comparison, we compare only with methods based on pretrained models, mostly BERT. We observe that BERT fine-tuned only with the source domain labeled data, without any knowledge of the target domain, is a competitive baseline. This source-only model even surpasses state-of-the-art methods developed for UDA, e.g. R-PERL \citep{ben2020perl}.

We reproduce the domain adversarial training procedure and present results in the DAT BERT column of Table~\ref{tab:results}. Adversarial training proved to be unstable in our experiments, even after careful tuning of the adversarial loss weighting factor \(\lambda_d\). This is evidenced by the high standard deviations in the DAT BERT experiments. We observe that adversarial training does not manage to outperform the source-only baseline.\footnote{Note that we did not have to perform extensive tuning for the other methods, including UDALM.}

Domain pretraining increases the average accuracy with an absolute improvement of \(0.85\%\) over the source-only baseline.
Continuing MLM pretraining on the target domain data leads to better model adaptation, and therefore improved performance, on the target domain.
This is consistent with previous works on supervised  \citep{gururangan-etal-2020-dont, xu-etal-2019-bert, sun2019fine} and unsupervised settings \citep{han-eisenstein-2019-unsupervised, du-etal-2020-adversarial}. 

UDALM yields an additional \(0.96\%\) absolute improvement of average accuracy over domain pretraining. Keeping the MLM loss during fine-tuning therefore, leads to better adaptation and acts as a regularizer that prevents the model from overfitting on the source domain. We also observe smaller standard deviations when using UDALM, which indicates that including the MLM loss during fine-tuning can result to more robust training.

\begin{figure*}[t]
\centering
\includegraphics[width=.8\textwidth]{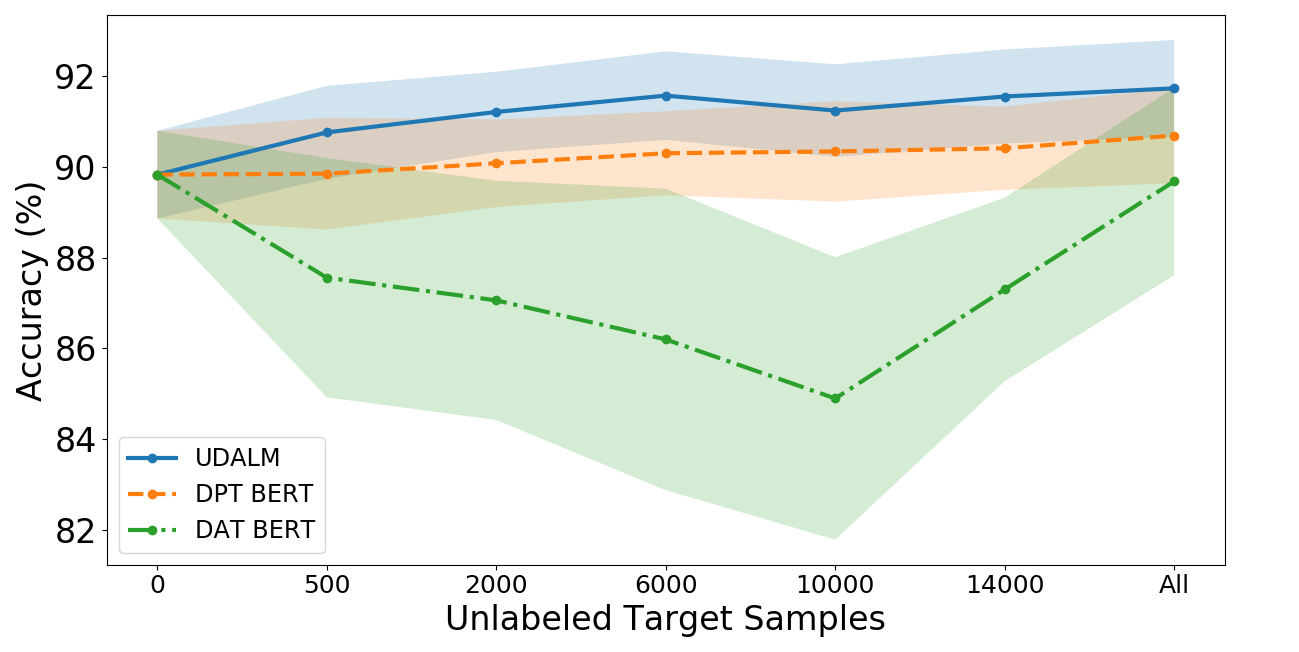}
\caption{Average accuracy for different amount of target domain unlabeled samples of: (1) DPT BERT (2) DAT BERT and (3) UDALM.}
\label{fig:efic}
\end{figure*}
\subsection{Sample efficiency}
\label{sec:sample-eff}

UDALM surpasses in terms of macro-average accuracy all other approaches for unsupervised domain adaptation on the Amazon reviews multi-domain sentiment dataset.
Specifically, our method improves on the state-of-the-art pseudo-labeling (p+CFd \citealp{ye_feature_2020}), domain adversarial (DAAT \citealp{du-etal-2020-adversarial}) and pivot-based (R-PERL \citealp{ben2020perl}) approaches by \(1.11\%\), \(1.62\%\) and \(4.24\%\) respectively.

We further investigate the impact of using different amount of target domain unlabeled data on model performance, to study the sample efficiency of UDALM. We experiment with settings of 500, 2000, 6000, 10000 and 14000 samples, by randomly limiting the number of unlabeled target domain data. For each setting we conduct three experiments with BERT models: (1) DPT, (2) DAT and (3) UDALM. When no target data are available, all methods are equivalent to a source only fine-tuned BERT. Again, we do not tune the hyper-parameters for DPT or UDALM. Fig.~\ref{fig:efic} shows the average accuracy on the twelve adaptation scenarios of the studied dataset. We see that UDALM produces robust performance improvement when we limit the amount of target data, indicating that it can be used in low-resource settings. However, training BERT in a domain adversarial manner shows instabilities. This is further discussed in Section~\ref{discussion}. 

\subsection{On the stopping criteria for UDA training}

\begin{table}[b]
\centering
\begin{tabular}{|l l c|}
\hline
Stopping Criterion & Epochs & Av. Acc. \\
\hline
\hline
Fixed  & 1 & 90.98  \\
Fixed  & 3 & 91.65 \\
Fixed  & 10 & \textbf{91.75} \\
Min source loss & 10, patience 3 & 91.30 \\
Min mixed loss & 10, patience 3 & \textbf{91.74} \\
\hline
\end{tabular}
\caption{Comparison of average accuracy for various validation settings.}
\label{validation}
\end{table}

% asto se mena auto

A common problem when performing UDA is the lack of target labeled data that can be used for hyperparameter validation. For example, \citet{ruder-plank-2018-strong} use a small set of labeled target data for validation, putting the problem in a semi-supervised setting. When training under a domain shift, optimization of model performance on the source data may not result to optimal performance for the target data. 

To illustrate this, we examine if the minimization of the mixed loss can be used as a stopping criterion for UDA training.
% We explore if validation can be performed on the proposed mixed loss. 
We compare five stopping criteria: (1) fixed training for 1 epoch, (2) fixed training for 3 epochs, (3) fixed training for 10 epochs, 
(4) stop when the minimum classification loss is reached for the source data and (5) stop when the minimum mixed loss ( Eq.~\ref{loss}) is reached. For (4) and (5) we train for 10 epochs with patience 3. 
We report average accuracy of the five stopping criteria over the twelve adaptation scenarios of Amazon Reviews dataset on Table~\ref{validation}. 
Training for a fixed number of 10 epochs and stopping when the minimum mixed loss perform best, yielding comparable accuracies of $91.75\%$ and $91.73\%$ respectively.
Note that stopping when the minimum source loss stops the fine-tuning process too soon and does not allow the model to learn the target domain effectively.
Overall, we observe that the mixed loss can be effectively used for early stopping, regularizing the model and alleviating the need for extensive search for the optimal number of training steps.
This is an indication that the mixed loss could be used for model validation.

\begin{figure*}[ht]
\centering
\includegraphics[width=0.7\textwidth]{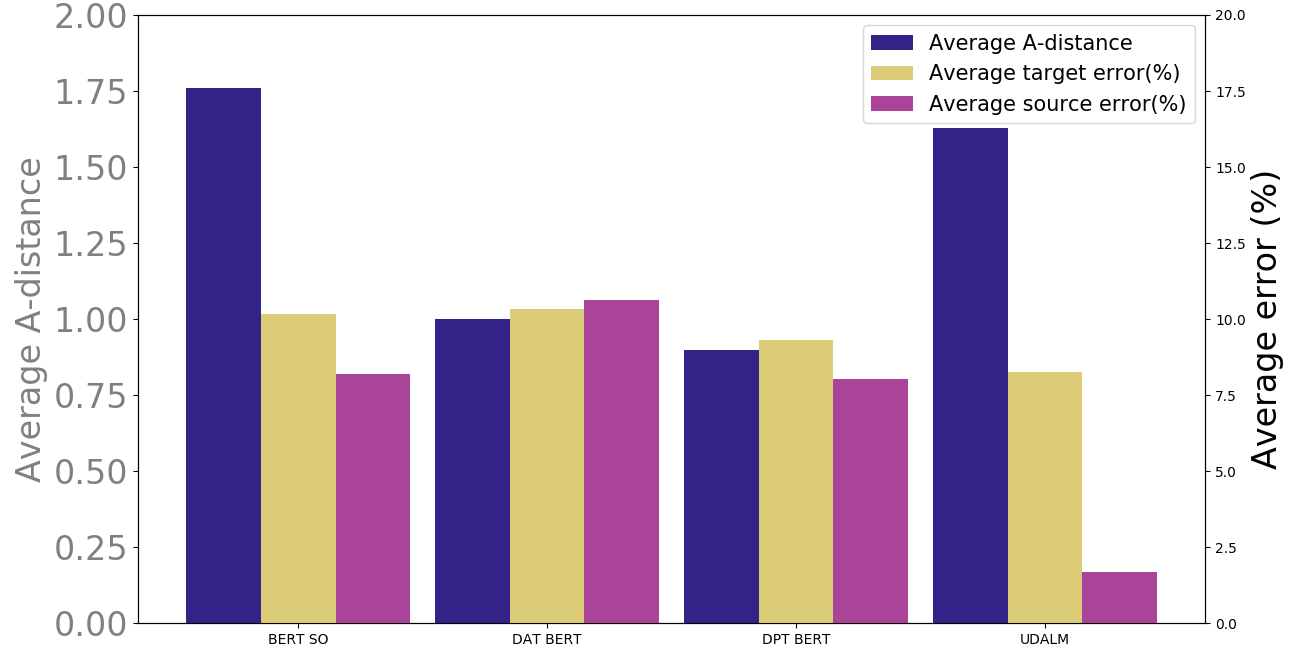}
\caption{Comparison of average A-distance, average source error and average target error rate of different methods over all source - target pairs of the Amazon reviews dataset. }
\label{fig:adistance}
\end{figure*}

\section{Discussion}
\label{discussion}
\subsection{Background Theory}

\citet{ben2007analysis, ben2010theory} provide a theory of learning from different domains.
A key outcome of this work is the following theorem:

\paragraph{Theorem}\citep{ben2007analysis, ben2010theory}
\label{theorem:paragraph}
 Let \(H\) be the hypothesis space and let \(D_S, D_T\) be the two domains and \(\epsilon_S, \epsilon_T\) be the corresponding error functions. Then for any \(h \in H\):

\begin{equation}
\label{theorem}
    \epsilon_T(h) \leq \epsilon_S(h) + \frac{1}{2} d_{H \Delta H}(D_S, D_T) + C
    % \min_{h'\in H}(\epsilon_T(h')+\epsilon_S(h'))
\end{equation}
where \(d_{H \Delta H}(D_S, D_T)\)  is the \(H \Delta H\)-divergence \citep{kifer2004detecting} between two domains, that is a measure of distance between domains that can be estimated from finite samples.

Eq.~\ref{theorem} defines an upper bound for the expected error \(\epsilon_T(h)\) of a hypothesis \(h\) on the target domain as the sum of three terms, namely the expected error on the source domain \(\epsilon_S(h)\), the divergence between the source and target domain distributions \(\frac{1}{2} d_{H \Delta H}(D_S, D_T)\) and the error of the ideal joint hypothesis \(C\). When such an hypothesis exists, the term is considered relatively small and in practice ignored.
The first term, bounds the expected error on the target domain by the expected error in the source domain and is expected to be small, due to supervised learning on the source domain. The second term, gives a notion of distance between the source and target domain extracted features. 
Intuitively this equation states: ``if there exists a hypothesis \(h\) that has small error on the source data and the source feature space is close to the target feature space, then this hypothesis will have low error on the target data''. Domain Adversarial Training aims to learn features that simultaneously result to low source error and low distance between target and source feature spaces based on the combined loss in Eq.~\ref{eq:datobj}.

\subsection{A-distance only provides an upper bound for target error}

According to \citet{ben2007analysis} the \(H \Delta H\)-divergence can be approximated by proxy A-distance, that is defined by Eq.~\ref{eq:adis} given the domain classification error \(\epsilon_{D}\). 
\begin{equation}
    \label{eq:adis}
    d_{A} = 2(1-2\epsilon_{D})
\end{equation}

We calculate an approximation of the distance between domains. Following prior work \citep{ganin2016domain, saito2017asymmetric} we create an SVM domain classifier. 
We feed the SVM with BERT's \([CLS]\) token representations, measure the domain classification error, and compute A-distance as in Eq.~\ref{eq:adis}. 
We train the domain classifier on 2000 samples from each source and target domains. 
Fig.~\ref{fig:adistance} shows the A-distance along with the source and the target error, averaged over the twelve available domain pairs using representations obtained from four methods, namely BERT SO, DAT BERT, DPT BERT and UDALM.
DAT BERT minimizes the distance between domains.
DPT BERT also reduces the A-distance, to similar levels with DAT, without using an explicit loss to minimize A-distance.
To our surprise we found that, although it achieves the lowest error rate, UDALM does not significantly reduce the proxy A-distance compared to the source-only baseline.
Additionally, we observe that the source error is correlated to model performance on the target task, i.e. models with lower source error have also lower target error.
UDALM specifically, achieves high accuracy on the source task and is able to transfer the task knowledge across domains, while DAT is able to bring domain representations closer, but at the cost of achieving weaker performance on the task at hand.

Overall, we do not observe a correlation between the resulting A-distance and model performance on target domain.
Therefore, lower distance between domains, achieved intentionally or not, is not a necessary condition for good performance on the target domain\footnote{\citet{shu2018dirt} state that feature distribution matching is a weak constraint when high-capacity feature extractors are used. Intuitively, a high-capacity feature extractor can perform arbitrary transformations to the input features in order to match the distributions.}, and our efforts could be better spent towards synergistic learning of the supervised source task and the target domain distribution.

% Additionally, optimizing performance on the target domain does not require minimizing the distance between domains.
% \footnote{The above is aligned with the results of \citet{shu2018dirt}, stating that, if the feature extraction function has high-capacity then domain adversarial training is not sufficient for domain adaptation.}

\begin{figure*}[ht]
\centering
\begin{subfigure}[t]{0.34\textwidth}
\centering
         \includegraphics[width=\textwidth]{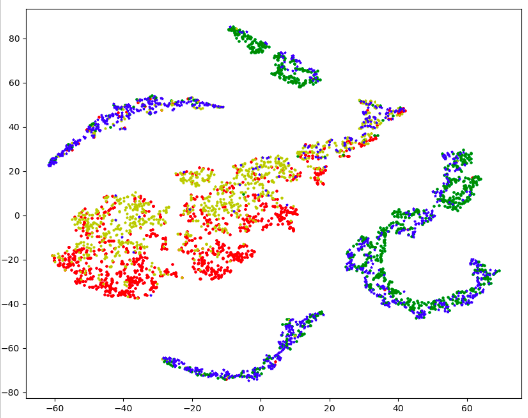}
         \caption{DAT BERT}
         \label{fig:advtsne}
\end{subfigure}
\begin{subfigure}[t]{0.34\textwidth}
\centering
         \includegraphics[width=\textwidth]{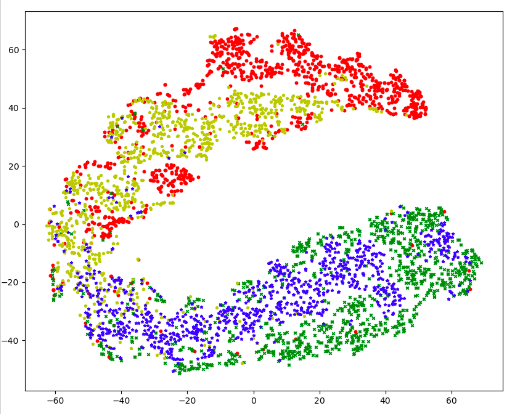}
         \caption{BERT SO}
         \label{fig:atsne}
\end{subfigure}

\begin{subfigure}[t]{0.34\textwidth}
\centering
         \includegraphics[width=\textwidth]{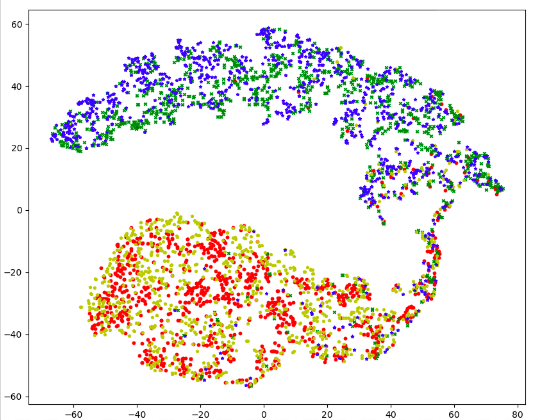}
         \caption{DPT BERT}
         \label{fig:btsne}
\end{subfigure}
\begin{subfigure}[t]{0.34\textwidth}
\centering
         \includegraphics[width=\textwidth]{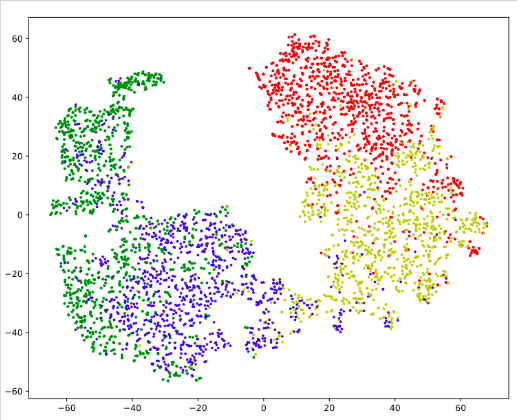}
         \caption{UDALM}
         \label{fig:ctsne}
\end{subfigure}
\begin{subfigure}[t]{0.6\textwidth}
\centering
         \includegraphics[width=\textwidth]{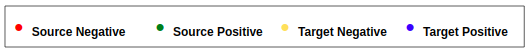}
         \label{fig:leg}
\end{subfigure}
\vspace*{-4mm}
\caption{$2D$ representations of  BERT \([CLS]\) features using t-SNE for the \(D \rightarrow K\) task.
%Red, green, yellow and blue dots correspond to source negative, source positive, target negative and target positive samples respectively. 
The goal is to maximize separation between target positive (blue) and target negative (yellow) samples.
}
\label{fig:tsne}
\end{figure*}

\subsection{Limitations of Domain Adversarial Training}

Domain adversarial training \citep{ganin2016domain} faces some critical limitations that make the method difficult to be reproduced due to high hyper-parameter sensitivity and instability during training. 

Such limitations have been highlighted by other authors in the UDA literature. For example, according to 
\citet{shen2017wasserstein} when a domain classifier can perfectly distinguish target from source representations, there will be a gradient vanishing problem. \citet{shah-etal-2018-adversarial} state that domain adversarial training is unstable and needs careful hyper-parameter tuning for their experiments. 
\citet{wang2020meta} report results over three multi-domain NLP datasets, where domain adversarial training in conjunction with BERT under-performs. \citet{ruder-plank-2018-strong} found that the domain adversarial loss did not help for their experiments on the Amazon reviews dataset.

In our experiments we note that domain-adversarial training results to worse performance than naive source only training. Furthermore, we experienced the need for extensive tuning of the \(\lambda_d\) parameter from Eq.~\ref{eq:datobj} every time the experimental setting changed (e.g. when testing for different amounts of available target data as in Section~\ref{sec:sample-eff}). This motivated us to further investigate the behavior of BERT fine-tuned with the adversarial cost.
For visual inspection, we perform T-SNE \citep{maaten2008visualizing} on representations extracted from BERT, under four UDA setings in Fig.~\ref{fig:tsne}. 
In Fig.~\ref{fig:advtsne} we observe features extracted using BERT with Domain Adversarial Training and we compare it with features from SO BERT (Fig.~\ref{fig:atsne}), DPT BERT (Fig.~\ref{fig:btsne}) and UDALM (Fig.~\ref{fig:ctsne}).
We observe that domain adversarial training manages to group tightly target and source samples, especially in the case of positive samples. Nevertheless, in the process, DAT introduces significant distortion in the semantic space, which is reflected in model performance\footnote{Note, we include this visualization for a single source-domain pair as an example. We performed multiple runs of T-SNE over all $12$ source-domain pairs and this behavior appeared consistently.}.

We can attribute this behavior to two factors. First, The formulation of the adversarial loss in Eq.~\eqref{eq:datobj} can lead to trivial solutions. In order to maximize the \(L_{ADV}\) term of  Eq.~\eqref{eq:datobj}, the model can just flip all domain labels, namely just predict that source samples belong to the target domain and vice-versa. In this case the model can still discriminate between domains and domain-independent representations are not encouraged. We empirically observed this behavior in our experiments with DAT, and only extensive hyper-parameter tuning could alleviate this issue.
Additionally, Eq.~\eqref{eq:datobj} aims to minimize the upper bound of the target error \(\epsilon_T(h)\) in Eq.~\eqref{theorem}. While this is desirable, reduction of the upper bound does not necessarily result in reduction of the bounded term in all scenarios.  
Furthermore, optimizing the \(L_{ADV}(\theta;D_S,D_T)\) term can lead to increasing \(L_{CLF}(\theta;D_S)\), and therefore one must find a balance between the two adversarial terms, again through careful hyper-parameter tuning.
These issues could potentially be alleviated by including regularization terms that discourage trivial solutions and improve robustness.
Therefore, given the lack of guarantees for good performance and the practical considerations, further investigation should be conducted regarding the robustness and reproducibility of DAT for UDA.

\section{Conclusions and Future Work}

Unsupervised domain adaptation of pretrained language models is a challenging problem with direct real world applications. 
In this work we propose UDALM, a robust, plug and play training strategy, which is able to improve performance in the target domain, achieving state-of-the-art results across $12$ adaptation settings in the multi-domain Amazon reviews dataset. 
Our method produces robust results with little hyper-parameter tuning and the proposed mixed-loss can be used for model validation, allowing for fast model development.
Furthermore, UDALM scales with the amount of available unsupervised data from the target domain, allowing for adaptation in low-resource settings.
In our analysis, we discuss the relationship between the A-distance and the target error. We observe that low A-distance may not suggest low target error for high capacity models. Additionally, we examine limitations of Domain Adversarial Training and highlight that the adversarial cost may lead to distortion of the feature space and negatively impact performance.
% We conclude this work by performing an discussion on the dominant Domain Adversarial training approach. We find that Domain Adversarial Training under-performs for our experiments, while it exhibits instability during training. We propose that this instability results from the introduction of the adversarial term and can be alleviated only with careful hyper-parameter selection, while the under-performance can be due to the lack of strict theoretical guarantees of this method.  

In the future we plan to apply UDALM to other tasks under domain-shift, such as sequence classification, question answering and part-of-speech tagging.
Furthermore, we plan to extend our method for temporal and style adaptation, by adding more relevant auxiliary tasks that model language shift over time and over different platforms. Finally, we want to investigate the effectiveness of the proposed fine-tuning approach in supervised scenarios.

\section*{Acknowledgements}
\begin{itemize}
    \item This research has been co‐financed by the European Regional Development Fund of the European Union and Greek national funds through the Operational Program Competitiveness, Entrepreneurship and Innovation, under the call RESEARCH – CREATE – INNOVATE (project safety4all with code:T1EDK-04248)
    \item This work has been partially supported by computational time granted from the Greek Research \& Technology Network (GR-NET) in the National HPC facility - ARIS.
    \item The authors would like to thank Efthymios Georgiou for his comments and suggestions.
\end{itemize}

\bibliographystyle{acl_natbib}
\bibliography{anthology,custom}

\end{document}